\documentclass[runningheads]{llncs}
\usepackage[T1]{fontenc}
\usepackage{graphicx}
\usepackage{float}
\usepackage{caption}
\usepackage{amsmath} 
\usepackage{subcaption}
\usepackage{xcolor}

\usepackage{bbding}
\begin{document}
\title{Towards Effective Long-Video Event Prediction via Multi-Level Event Semantics Mining}
% \author{MMM Anonymous Author(s)}
\author{Bo Peng \and
YuanJie Lyu \and
PengGang Qin \and Tong Xu\inst{(}\Envelope\inst{)}}

\institute{University of Science and Technology of China\\
\email{tongxu@ustc.edu.cn}}
% \email{tongxu@ustc.edu.cn}} 
% , USA \and
\maketitle              % typeset the header of the contribution
\begin{abstract}
Accurately predicting future events is fundamental to content understanding and decision-making across  various domains. While prior research has primarily focused on text or short-video scenarios, long-video event prediction—characterized by vast multimodal context and more complex narratives—remains underexplored. Meanwhile, although recent Long-Video Language Models (LVLMs), built on Large Language Models (LLMs) and Vision–Language Models (VLMs), have shown promise in long-video question answering and summarization,  they struggle to generalize to event prediction, as they can neither precisely extract event-related details nor perform fine-grained analysis of event development. To address this gap, we propose VISTA, a multi-level event semantics mining framework for long-video event prediction. Initially, VISTA applies a character-centric visual prompt to precise extract event-related visual details, enhancing detail-level semantics; subsequently, it employs a knowledge-enhanced iterative retrieval strategy, guiding the LLM to progressively construct logically coherent event chains, thereby improving event-level narratives; ultimately, VISTA adopts a human-like propose-then-retrieve strategy to generate diverse future-oriented proposals and integrate multi-level clues, producing robust and accurate predictions. Extensive experiments on real-world datasets validate the effectiveness of VISTA for long-video event prediction.

\end{abstract}
%
%
%

% 多尺度

\section{Introduction}
Future event prediction (FEP) aims to forecast the future trend of events based on historical information, thus providing strong support for content understanding and decision-making. Traditional FEP researches focus on the textual descriptions of historical events~\cite{halawi2024approaching,li2021future,ma2023context}
or short videos that typically span only a few seconds~\cite{lai2024object,lei2020more}. However, with the development of multimedia technologies, more and more events are presented in long videos like movies and TV series.
The complexity of long videos—characterized by vast multimodal contexts and intricate event narratives—poses significant challenges that existing FEP methods fail to address.
\vspace{-0.7cm}
\begin{figure}[H]
    \centering
    \includegraphics[width=\linewidth]{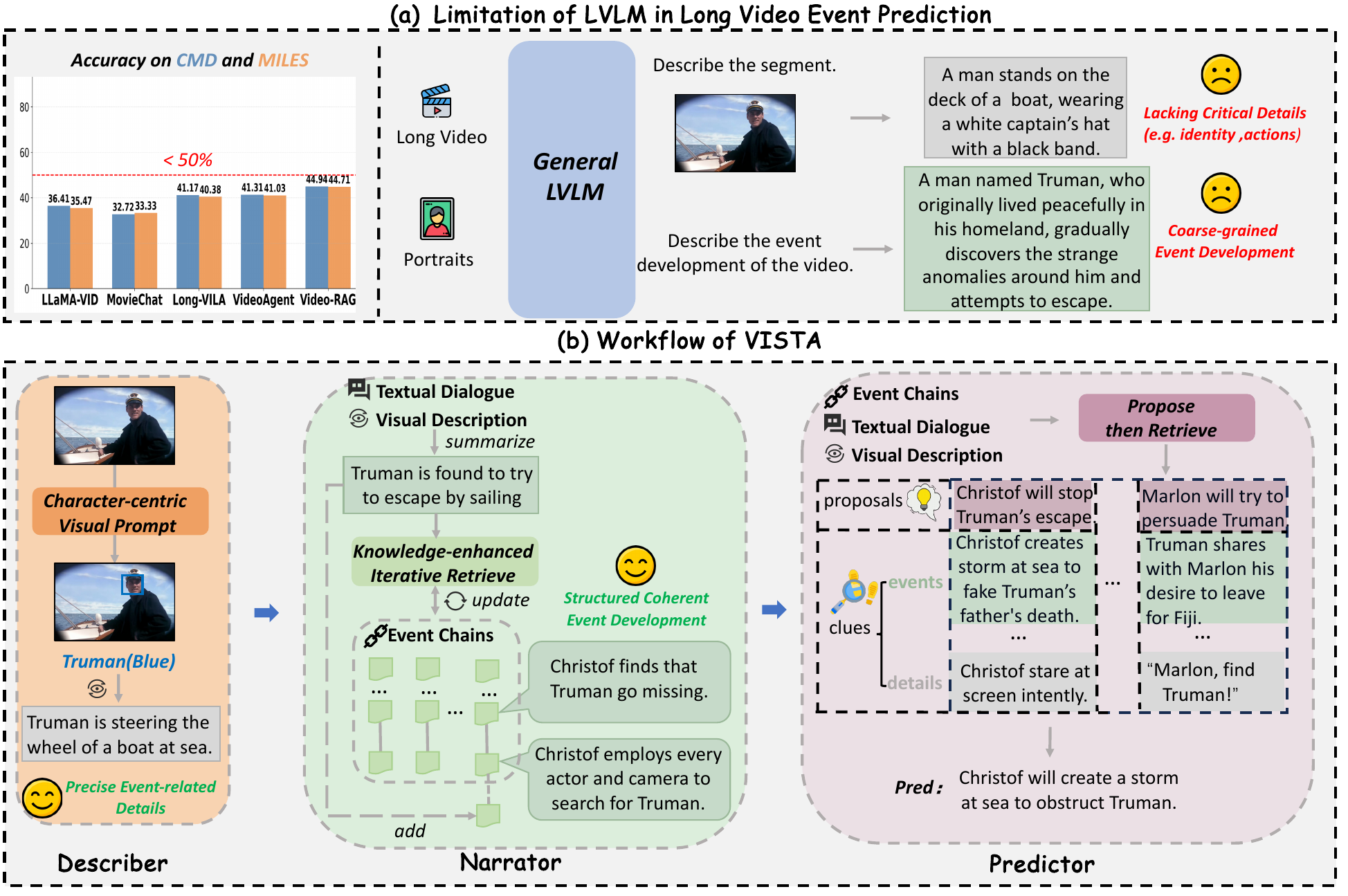}
    \caption{ Illustration of the limitation of general LVLMs on long-video event prediction, and the overall workflow of VISTA. }
    \label{fig:fig1}
\end{figure}
\vspace{-0.7cm}

Fortunately, the recent emergence of Large Language Models (LLMs) and Vision-Language Models (VLMs) has demonstrated strong capabilities in content understanding and reasoning, shedding light on the effective analysis of long-video content. Building on these advances, multiple Long-Video Language Models (LVLMs)\cite{fei2024video,luo2024video,li2025llama,song2024moviechat} have been developed, showing promising performance on long-video question answering and summarization tasks. However, as illustrated in Fig.\ref{fig:fig1}(a), their poor performance on long-video event prediction suggests that effectively leveraging LLMs and VLMs for this task is far from trivial, primarily due to the following critical issues:

\begin{enumerate}
    \item[\textbullet] \textbf{Semantic limitation of visual details.} Current VLMs focus more on  low-level visual elements~\cite{he2024storyteller} (e.g. objects, costumes) than on high-level event-related cues (e.g. character identity and actions). This bias introduces interference from irrelevant details and omits crucial event-related information, thereby undermining the detail-level event semantics of long videos.
    
    \item[\textbullet] \textbf{Coarse-grained reflection of event development.} Although LVLMs can summarize long videos, such summaries are  highly conclusive,  while lacking fine-grained analysis of event development. The resulting obscured event development logic and key events omission lead to substantial event-level semantic loss, and further affect future-oriented event reasoning.
    % LVLM可以提供长视频的粗粒度概述，却lack fine-grained analysis of the logic and causal relationships of event development
\end{enumerate}

To address these challenges, we propose VISTA, a multi-level event semantics mining framework for long-video event prediction. VISTA progressively distills detail- and event-level semantics and integrates multi-level cues to enable robust and reliable prediction. As shown in Fig.\ref{fig:fig1}(b), it consists of three modules:

Initially, the \textbf{Describer} apply \textbf{\textit{character-centric visual prompts}}, constructed through fine-grained in-frame feature matching, to raw video frames, to focus the VLM on character identities and actions, which serve as the foundational visual semantics of events.

Subsequently, the  \textbf{Narrator} integrates these visual details with transcribed dialogues into textual event descriptions and applies a \textbf{\textit{knowledge-enhanced iterative retrieval}} strategy, where a commonsense expert aligns causal relations between historical and current events, to progressively integrate them into structured event chains. Each chain forms a logically coherent sequence of events, enabling fine-grained and coherent event-level narrative. 

Ultimately, the \textbf{Predictor} employs a \textbf{\textit{propose-then-retrieve}} strategy to extend and integrate the multi-level event semantics established above. Inspired by human cognition, this strategy mirrors how people first generate multiple hypotheses from past event trajectories and then retrieve pertinent cues to refine the most plausible one. Concretely, the \textbf{Predictor} extends each event chain to produce diverse, future-oriented proposals and, via multi-granular feature matching, retrieves clues at both the detail- and event-levels for each proposal. The combination of coarse-grained proposals with fine-grained, multi-level clues enables the LLM to deliver reliable, robust event predictions.

The technical contributions of this paper are as follows:
\begin{enumerate}
\item[\textbullet] To the best of our knowledge,  we are among the first ones who elaborate on predicting future events based on long videos.
\item[\textbullet] We propose VISTA, a multi-level event semantics framework that: (i) extracts precise event-related visual details, (ii) constructs fine-grained and coherent event chains, and (iii) adopts a human-like propose-then-retrieve strategy to fully leverage multi-level event semantics for reliable prediction.
\item[\textbullet] Extensive experiments on real-world datasets demonstrate the superiority of VISTA, while ablation studies validate the effectiveness of each module.
\end{enumerate}

\section{Related Work}
\textit{\textbf{Future Event Prediction}} Early studies on future event prediction primarily focused on well-structured events such as texts from ontology and news articles~\cite{halawi2024approaching,li2021future,ma2023context}. They generally adopted graph neural networks to model structured events and formulated future event prediction as a link prediction task on graph networks. With the development of multimedia technologies, research has been extended to short videos. \cite{lei2020more} first introduced the video event prediction task and propose a transformer-based method to integrate multimodal clues from video content. \cite{lai2024object} advanced the field by integrating an object-centric cross-modal reasoning chains into VLMs. While these methods are effective for local event reasoning within short videos, they cannot handle the vast multimodal information of long videos, while lacking sufficient reasoning ability to analyze long-horizon event development.

\noindent \textit{\textbf{Long-Video-Language Models}}\label{subsec:lvu}
The strong understanding and reasoning capabilities of Large Language Models (LLMs) and Vision-Language Models (VLMs) have driven significant advances in multimodal research. Building on them, various general Long-Video Language Models (LVLMs) have emerged and shown effectiveness in long video understanding. Structurally, these models fall into two categories: end-to-end models~\cite{chen2024longvila,song2024moviechat,li2025llama,luo2024video,song2024moviellm} and agent-based models~\cite{ataallah2024goldfish,fan2025videoagent,luo2024video,wang2025videoagent}. End-to-end models aim to manage the vast multimodal context of long videos through efficient token-level representation or memory mechanisms. Agent-based models, in contrast, leverage LLMs as central agents in combination with VLMs for visual perception, enabling interactive extraction and aggregation of key information. Although effective for long-video question answering and summarization, we find they fail to effectively capture and reason over event development in long videos. In this work, we address this gap by enhancing multi-level event semantic mining and predictive analysis based on LLMs and VLMs.

\begin{figure*}[h]
  \centering
  \includegraphics[width=1\textwidth]{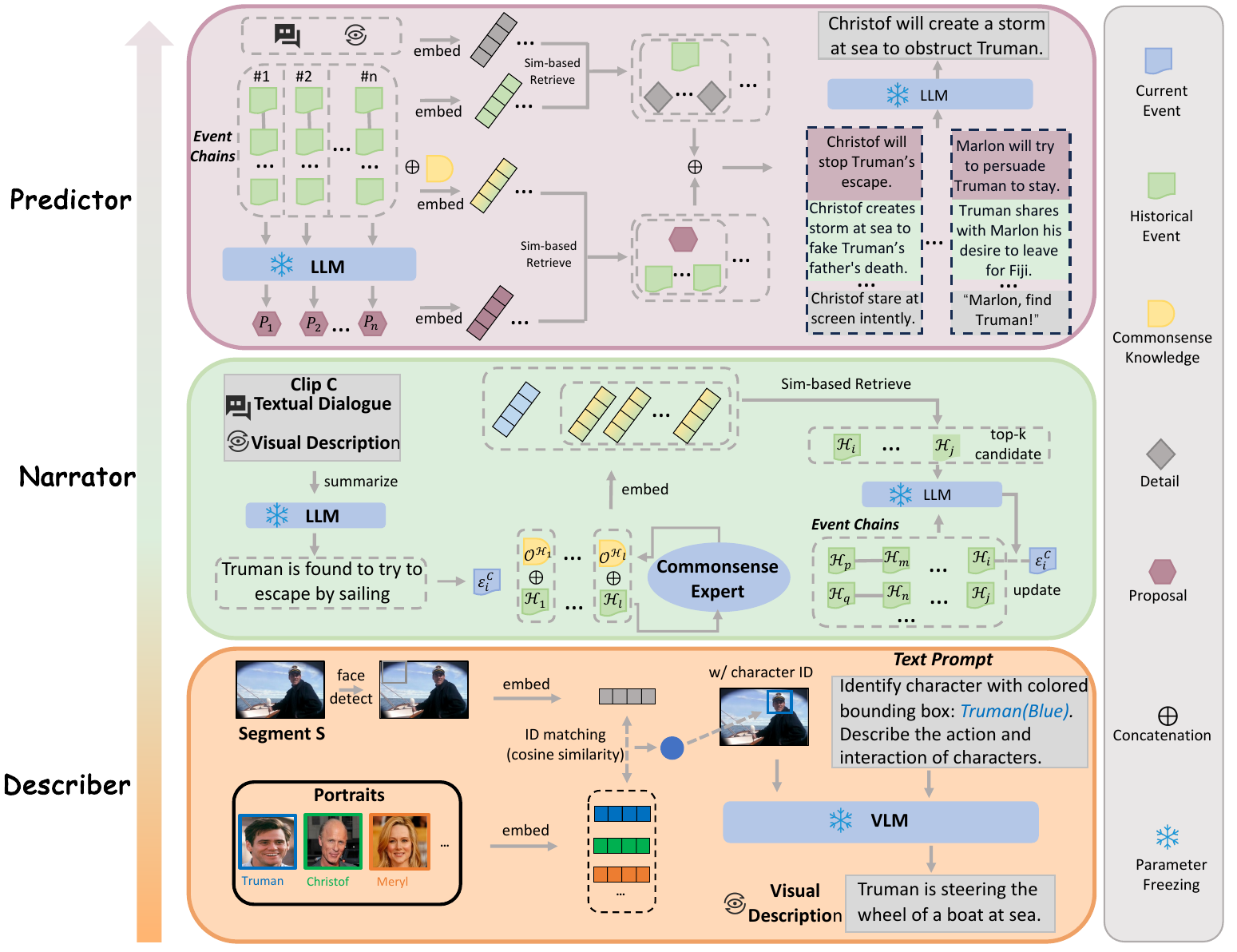}
  \caption{Illustration of our VISTA framework: \textcolor[HTML]{FFD55D}{Describer} featured by \textbf{\textit{character-centric visual prompt}} for more precise event-related visual detail extraction, \textcolor[HTML]{B0D59B}{Narrator} featured by \textbf{\textit{knowledge-enhanced iterative retrieval}} to generate coherent event chains, and \textcolor[HTML]{C7A5B0}{Predictor} featured by \textbf{\textit{propose-then-retrieve}} strategy to effectively utilize detail- and event-level cues for more precise prediction.}
  \label{figure:framework}
\end{figure*}

\section{Methodology}
We now introduce VISTA, a multi-level event semantics framework for long-video event prediction. Fig.\ref{figure:framework} illustrate the key modules of VISTA.

\subsection{Problem Formulation} \label{subsec:problem-formulation}
We formally define the long-video event prediction task as follows:
\begin{enumerate}
\item[\textbullet]  \textbf{Input}: A raw long video $V$ spanning tens of minutes along with portraits of relevant characters $\{P_i\}_{i=1}^K$.
\item[\textbullet] \textbf{Output}: A textual description of predicted future event $\hat{\mathcal{E}}$.
\end{enumerate}
% It should be noted that previous work~\cite{he2024storyteller, han2024autoad} has demonstrated that character portraits are essential prior knowledge for long video narratives. Following them, we define character portraits as a part of the input.

\subsection{Multimodal Preprocessing} \label{subsec:multimodal-preprocessing}
Given the raw long video $V$, we begin by utilizing WhisperX~\cite{bain2023whisperx}, a widely used ASR tool, to transcribe the audio of the long video into textual dialogue $T$, which serve as valuable clues for event understanding and reasoning.

Next, we used PySceneDetect, an established scene detection tool, to identify scene variations. The original video $V$ is then divided into shorter video segments $\{S_i\}_{i=1}^N$ based on on the detected scene boundaries. Considering that overly short segments could lead to insufficient context, we further concatenate consecutive segments into video clips $\{C_i\}_{i=1}^M$ in chronological order, ensuring each clip spans no less than 3 minutes.

\subsection{Describer: Character-centric Visual Prompt} \label{subsec:cvd}
Prior works~\cite{shtedritski2023does} have proven that visual prompts can effectively highlight key visual elements to facilitate more precise visual clue extraction of VLMs. Motivated by this, we propose character-centric visual prompts to focus VLM on character identity and actions—fundamental semantics of events.

Given a video segment $S$, we first employ InsightFace~\cite{ren2023pbidr}, an off-the-shelf face detection model, to detect facial regions frame-by-frame in the form of bounding boxes. We then utilize CLIP~\cite{radford2021learning} to extract facial embeddings from these bounding boxes as $\{f_i\}_{i=1}^N$. Similarly, each character portrait $P_i$ is processed by CLIP, forming a set of portrait embeddings $\{p_i\}_{i=1}^K$. Subsequently, we compute the cosine similarities between the two sets of embeddings and identify the best-matching portrait face for each face bounding box. Formally,
\begin{equation}
bbox_j = \arg\max_{i \in \{1, ..., K\}} \left( \frac{p_i \cdot f_j}{|p_i||f_j|} \right)
\end{equation}

We overlay each bounding boxes with colors corresponding to its best-matching characters, where each character is pre-assigned a unique color. The video frames with colored bounding boxes are then fed into the VLM, accompanied by a text prompt that explicitly specifies the color-to-character mappin, (e.g. \textit{Truman(Blue)}). This visual-text collaborative prompt emphasizes character identity and action to VLM, thereby enhancing accurate event-related visual details:
\begin{equation}
vd = \mathcal{VLM}(S; \{bbox_j\}_{j=1}^N)
\end{equation}

\subsection{Narrator: Knowledge-enhanced Iterative Retrieve}
Leveraging the powerful summarization capabilities of LLMs, we can reliably summarize the transcribed dialogues $D_C$ and visual descriptions $vd_C$ within the temporal domain of each video clip $C$, yielding chronologically ordered event descriptions $\{\mathcal{E}^C_j\}_{j=1}^l$, where $\mathcal{E}^C_j$ corresponds to the $j$-th event occuring in $C$:

\begin{equation}
\{\mathcal{E}^C_j\}_{j=1}^l = \mathcal{LLM}(D_C, vd_C)
\end{equation}

\begin{figure}[H]
    \centering
    \includegraphics[width=\linewidth]{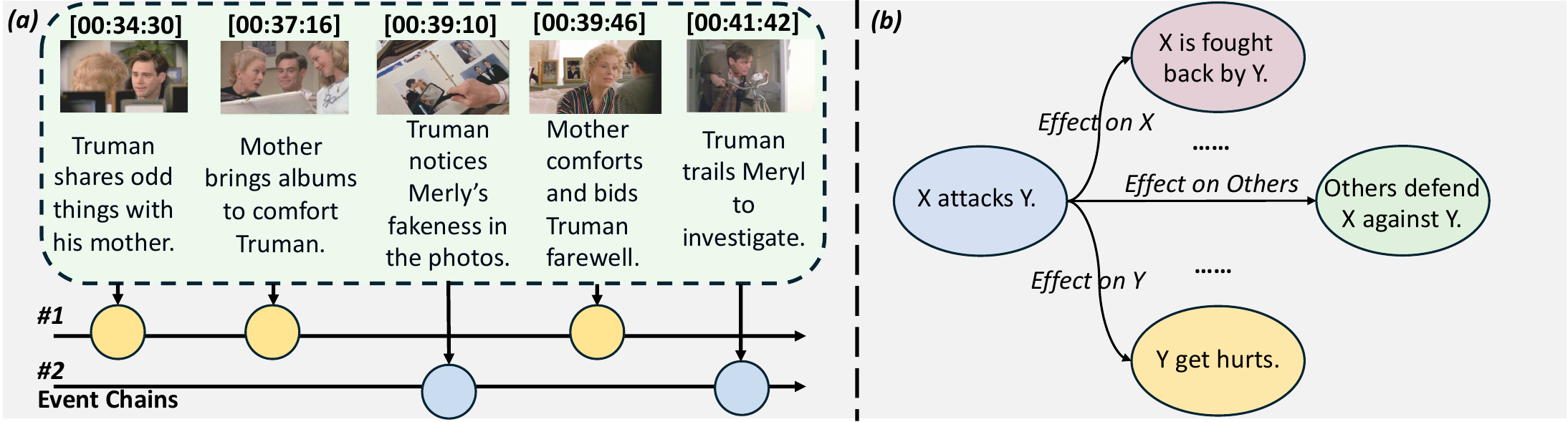}
    \caption{Illustration of event chains and commonsense expert.}
    \label{fig:event}
\end{figure}

However, as illustrated in the Fig.\ref{fig:event}, temporally sequential events in long video narratives often lack logical coherence, potentially hampering subsequent reasoning and prediction. To address this, we further integrate events into structured event chains—each reflecting clear causality—using a knowledge-enhanced iterative retrieval strategy, to produce logically coherent event-level narratives.
  
This strategy processes events iteratively in chronological order. In each iteration, the event being processed is treated as the current event $\mathcal{E}$, and all preceding events are considered historical events $\{\mathcal{H}_i\}_{i=1}^h$. To strengthen the correlation between causally related events for more accurate identification, we introduce an off-shelf LLM fine-tuned with a large-scale commonsense knowledge graph as a commonsense expert \( f_{CE} \) to generate comprehensive  potential causal outcomes $\{\mathcal{O}^{\mathcal{H}_i}_j\}_{j=1}^o$ of each $\mathcal{H}_i$, as shown in Fig.~\ref{fig:event}. These causal commonsense insights have been shown by previous research~\cite{lyu2024generating} to effectively highlight the semantic similarity between causally related events. Thus, we embed the concatenated pairs of each historical event with its causal outcomes, as well as the current event $\mathcal{E}$, using an embedding model $f_t$. The embedding of the current event $\mathbf{e}^{\mathcal{E}}$ is set as the pivot, and the cosine similarity between it and each historical event embedding $\mathbf{e}^{\mathcal{H}_i}_j$ is calculated accordingly:

\begin{equation}
\text{sim}(\mathbf{e}^{\mathcal{H}_i}_j,  \mathbf{e}^{\mathcal{E}}) = \frac{\mathbf{e}^{\mathcal{H}_i}_j \cdot \mathbf{e}^{\mathcal{E}}}{\|\mathbf{e}^{\mathcal{H}_i}_j\| \|\mathbf{e}^{\mathcal{E}}\|}, \quad \mathbf{e}^{\mathcal{H}_i}_j = f_t(\mathcal{H}_i \oplus \mathcal{O}^{\mathcal{H}_i}_j), \quad \mathbf{e}^{\mathcal{E}} = f_t(\mathcal{E})
\end{equation}

Based on the similarity scores, we select the top-
top-\(k\) historical events as the final candidates, as a higher similarity indicates a greater potential for causality. Top-\(k\)(\(A\), \(f\)) denotes selecting the top \(k\) elements from the set \(A\), ranked in descending order of their \(f\) values:
\begin{equation}
\mathcal{H}_c = \text{Top-}k\left(\bigcup_{i=1}^l \{(\mathcal{H}_i, \mathcal{O}^{\mathcal{H}_i}_j) \mid j = 1, 2, \ldots \}, \text{sim}(e^{\mathcal{H}_i}_j, e^{\mathcal{E}})\right)
\end{equation}

The event chain each candidate locates, along with the causal guidance and the current event, are fed into the LLM.
The refined context and explicit causal guidance enable the LLM to effectively determine the event chain to which the current event should be added. Starting from the first event of the initial clip, the event chain is iteratively updated until the final one, ultimately forming a complete set of event chains that cover the entire video content.

\subsection{Predictor: Propose-then-Retrieve} \label{subsec:sp}
The details extracted by the Describer, along with the structured and coherent event chains constructed by the Narrator, establish a solid semantic foundation for event prediction. To fully leverage these multi-level clues, we design a propose-then-retrieve strategy that emulates the human cognition.

We begin by generating a diverse set of future proposals to capture a broad range of future potentials. Intuitively, each event chain $EC$ can be viewed as a branch of the storyline, inherently capable of extending into the future. Inspired by this, we feed each event chain into the LLM to generate multiple predictive proposals $\{\mathcal{P}_k\}_{k=1}^p$ that extend the storyline:
\begin{equation}
    \{\mathcal{P}_k\}_{k=1}^p = \mathcal{LLM}(EC)
\end{equation}

Subsequently, we retrieve relevant clues from multi-level historical information for each proposal ${\mathcal{P}_k}$. These clues help highlight and refine the most probable proposal into precise prediction. Specifically, we first reuse the embeddings of each event-outcome pairs $\mathbf{e}^{\mathcal{H}_i}_j$ to compute cosine similarity with each proposal embedding $\mathbf{e}^{\mathcal{P}_k}$, with a threshold $\tau_e$ for filtering, to get event-level clues $\mathcal{C}^{\mathcal{E}}_{\mathcal{P}_k}$:

\begin{equation}
    \mathcal{C}^{\mathcal{E}}_{\mathcal{P}_k} = \{\mathcal{H}_i \mid  sim(\mathbf{e}^{\mathcal{H}_i}_j, \mathbf{e}^{\mathcal{P}_k}) > \tau_e, \forall i \in [1, |\mathcal{H}|], j \in [1, |\mathcal{O}^{\mathcal{H}_i}|]\}
\end{equation}

For more fine-grained detail-level clues, we further filter the detail-level clues within the temporal scope of each event-level clue. Considering that there is no causal semantic gap between events and their associated details, we directly compute the cosine similarity between the the raw embedding of each event and all details it contains , with a threshold $\tau_d$ for filtering, to get details-level clues $\mathcal{C}^{\mathcal{D}}_{\mathcal{P}_k}$.

Finally, we combine each proposal $\mathcal{P}_k$ with corresponding multi-granularity clues $\mathcal{C}^{\mathcal{E}}_{\mathcal{P}_k}$ and $\mathcal{C}^{\mathcal{D}}_{\mathcal{P}_k}$, and feed them  into an LLM to determine and refine the most probable proposal, based on its retrieved clues, into the final prediction $\hat{\mathcal{E}}$:
\begin{equation}
    \hat{\mathcal{E}} = \mathcal{LLM}(\{(\mathcal{P}_k, \mathcal{C}^{\mathcal{E}}_{\mathcal{P}_k}, \mathcal{C}^{\mathcal{D}}_{\mathcal{P}_k})\}_{k=1}^p)
\end{equation}

\section{Experiment}
\subsection{Experimental Settings}
\noindent \textbf{Dataset Description}
\noindent We adopt the following two evaluation datasets for long-video event prediction: (1) \textbf{CMD}: Adapted from the Condensed Movies (CMD) test set~\cite{bain2020condensed}. Original CMD collects abundant movies with timestamped textual event description annotations, each event presenting a key stage of the storyline. By segmenting the original movies according to the timestamp, we construct event-video pairs as evaluation samples. To prevent data leakage, following~\cite{ghermi2024short}, we feed anonymized movie synopses to every LLMs/VLMs used in our experiments to predict future events; any correctly identified movies are excluded. This results in 4,987 samples with an average duration of 48.6 minutes. (2)\textbf{MILES}: Contains 624 event-video pairs with an average duration of 78.5 minutes from 80 TV series released after January 2025, ensuring no overlap with the pretraining data of any models in the experiments. Annotation follows a two-stage process: two annotators select key events from each TV series, and two additional annotators independently review these selections to minimize bias and ensure quality. For both datasets, we include IMDb character portraits as prior knowledge.

\noindent \textbf{Evaluation Metrics.}
To comprehensively assess the performance of different models, we conducted both generative and multiple-choice evaluations. For the generative evaluation, we use widely adopted metrics in generation tasks: ROUGE-L(R)~\cite{lin2004rouge}, METEOR(M)~\cite{banerjee2005meteor}, and CIDEr(C)~\cite{vedantam2015cider}, along with Sentence-BERT (SB)~\cite{reimers2019sentence}. For the multiple-choice evaluation, we use accuracy (Acc).

\noindent \textbf{Compared Methods.} We selected various state-of-the-art general LVLMs for comprehensive comparison. As introduced in Section~\ref{subsec:lvu},they are grouped into two types: \textit{end-to-end models}, including MovieChat~\cite{song2024moviechat}, Llama-VID~\cite{li2025llama}, Movie-LLM~\cite{song2024moviellm}, 
LongVILA~\cite{chen2024longvila}, Video-CCAM~\cite{fei2024video}, LongLLaVA~\cite{wang2024longllava}, Video-XL~\cite{shu2025video} and \textit{agent-based models}, including VideoAgent (F)~\cite{fan2025videoagent}, VideoAgent (W)~\cite{wang2025videoagent} , Goldfish~\cite{ataallah2024goldfish}, Video-RAG~\cite{luo2024video}. For more comprehensive evaluation, we additionally fine-tune several end-to-end models on the CMD validation set, which contains 3,348 samples from 358 movies. Moreover, we include two state-of-the-art short video prediction methods—VLEP~\cite{lei2020more} and OC-2-Reasoning~\cite{lai2024object}—as baselines. Since these methods cannot process long video inputs, we feed them only the last 10 seconds of each video. Notably, VLEP lacks generative capability and is therefore evaluated solely under the multiple-choice setting.

\noindent \textbf{Implementation Details.} Our VISTA implementation adopts LLaMA-3.1-8B-Instruct\cite{grattafiori2024llama} as the default LLM and Qwen2.5-VL-7B-Instruct as the default VLM, with all-MiniLM-L6-v2\cite{reimers2019sentence} used as the embedding model across all components. The commonsense expert is implemented as a Flan-T5\cite{chung2024scaling} model fine-tuned on ATOMIC\cite{sap2019atomic}, following~\cite{lyu2024generating}. For the multiple-choice evaluation, GPT-4.1 is employed to generate three distractor options from the original event descriptions. The default LLM then selects the option most consistent with VISTA’s generated prediction as its answer. To prevent data leakage, all dialogue text and character names are anonymized throughout the evaluation. Relevant prompts of VISTA can be found in \url{https://anonymous.4open.science/r/VISTA-6BBF}.

\subsection{Performance Comparison.}
\vspace{-0.65cm}
\begin{table*}[!htbp]
\centering
\caption{Comparison of different methods on CMD and MILES datasets.}
\resizebox{\textwidth}{!}{%
\large
\begin{tabular}{c|cccc|c|cccc|c}
\hline
\multicolumn{1}{c|}{}         & \multicolumn{5}{c|}{\textbf{CMD}} & \multicolumn{5}{c}{\textbf{MILES}} \\ 

\cline{2-11}                                             & \textbf{R} & \textbf{M} & \textbf{C} & \textbf{SB} & \textbf{Acc(\%)} & \textbf{R} & \textbf{M} & \textbf{C} & \textbf{SB} & \textbf{Acc(\%)} \\
\hline\hline
MovieChat                       & 12.34 & 13.52 & 11.81 & 34.61 & 32.72 & 12.12 & 13.83 & 11.62 & 35.13 & 33.33   \\
Movie-LLM                      & 12.89 & 14.43 & 12.71 & 35.61 & 33.23 & 13.92 & 16.12 & 13.35 & 39.57 & 37.50  \\
LLaMA-VID                   & 12.54 & 14.57 & 12.75 & 36.36& 36.14 & 12.75 & 14.51 & 12.21 & 35.85 & 35.47 \\
LongVILA                & 14.72 & 18.23 & 15.31 & 42.61 & 41.17 & 14.41 & 17.71 & 15.01 & 42.32 & 40.38  \\
Video-CCAM             & 13.61 & 16.54 & 14.64 & 41.42 & 39.56 & 13.12 & 16.23 & 14.41 & 40.24 & 38.46 \\
LongLLaVA            & 13.25 & 16.71 & 14.34 & 39.72 & 39.32 & 13.44 & 16.41 & 14.79 & 39.71 & 38.61 \\
Video-XL            & 13.21 & 15.54 & 13.34 & 38.14 & 37.24 & 12.63 & 15.23 & 12.97 & 38.01 & 36.24 \\
\hline
MovieChat (ft)                       & 13.34 & 14.32 & 12.83 & 35.87 & 35.43 & 12.67 & 14.53 & 12.41 & 35.18 & 34.78   \\
LLaMA-VID (ft)                   & 12.91 & 14.83 & 13.32 & 37.64 & 37.02 & 12.87 & 14.35 & 12.48 & 36.85 & 35.10 \\
LongVILA (ft)                & 15.72 & 17.79 & 16.67 & 42.23 & 42.94 & 15.84 & 18.03 & 16.31 & 41.32 & 42.21  \\
Video-CCAM (ft)            & 16.23 & 18.54 & 16.64 & 41.42 & 42.31 & 15.72 & 18.51 & 16.21 & 40.24 & 41.03 \\
Video-CCAM (ft)            & 16.23 & 18.54 & 16.64 & 41.42 & 42.31 & 15.72 & 18.51 & 16.21 & 40.24 & 41.03 \\

\hline
VideoAgent (F)                  & 16.72 & 20.02 & 17.21 & 44.91 & 42.73 & 16.41 & 19.77 & 17.42 & 44.15 & 43.24  \\
VideoAgent (W)                  & 15.49 & 18.24 & 16.64 & 42.42 & 41.31 & 15.67 & 18.12 & 16.35 & 42.31 & 41.03 \\
Goldfish                          & 17.34 & 20.03 & 16.67 & 43.91 & 43.41 & 15.47 & 19.31 & 16.51 & 43.57 & 41.72  \\
Video-RAG                       & 17.78 & 22.06 & 18.33 & 46.06 & 44.94 & 17.78 & 21.71 & 18.13 & 45.62 & 44.71  \\
\hline
VLEP                             & - & - & - & - & 23.88 & - & - & - & - & 25.16 \\
OC-2-Reasoning                   & 7.12 & 6.24 & 5.45 & 30.18 & 25.64 & 7.16 & 8.14 & 6.93 & 32.14 & 27.88 \\
\hline
\textbf{VISTA (Ours)}               & \textbf{25.92} & \textbf{33.12} & \textbf{24.94} & \textbf{63.14} & \textbf{62.43} & \textbf{25.73} & \textbf{32.52} & \textbf{25.34} & \textbf{63.42} & \textbf{62.81}   \\
\hline
\end{tabular}
}
\label{table:main}
\end{table*}
\vspace{-0.5cm}
\noindent Table \ref{table:main} shows the performance comparison between VISTA and existing methods on the CMD and MILES, from which we can draw the following conclusions:

\begin{enumerate}
    \item [\textbullet] VISTA significantly outperforms all the other methods across both datasets, achieving absolute improvements of at least 8.14\%, 11.06\%, 6.61\%, 17.08\%, and 17.49\% across every metric in CMD, and at least 7.95\%, 10.81\%, 7.21\%, 17.80\%, and 18.10\% in MILES. The great performance exhibited by VISTA demonstrates its remarkable effectiveness for long video event prediction. 

    \item [\textbullet] On both CMD (movie-based) and MILES (TV-series-based), VISTA achieves stable performance with accuracy exceeding 60\%. This demonstrates its strong generality across different long-video genre, from movies featured by compact storytelling to TV series featured by loosely structured plots.

    \item [\textbullet]  End-to-end LVLMs perform poorly, and further task-specific fine-tuning only brings marginal performance gains, indicating that their end-to-end architectures fail to effectively extract and exploit event semantics for prediction.
    
    \item [\textbullet] Agent-based LVLMs generally outperform end-to-end models, benefiting from LLM-based retrieval and reasoning modules that improve fine-grained content analysis. However, their mediocre absolute performance suggests that such mechanisms remain insufficient to comprehensively capture and leverage multi-level event semantics.

    \item [\textbullet] Short video event prediction methods fail almost completely, with their accuracy approaching random prediction (25\%). They can neither handle the excessive multimodal information in long videos, nor predict high-level future events simply based on low-level reasoning of local details.
    
\end{enumerate}

\subsection{Ablation Study.}
\noindent \textbf{Effect of Multi-Level Semantics.}We compare VISTA with the following two variants to examine the role of detail- and event-level semantics plays in long-video event prediction, as shown in Table~\ref{tab:one}:
(1) \textit{\textbf{w/o VD}}: Without the visual details (VD), we only take textual dialogues as detail-level information, which leads to catastrophic performance loss, highlighting the indispensable event-related semantics visual details contain for long-video event prediction.
(2) \textit{\textbf{w/o EC}}: Without the event chains (EC) as event-level narratives, we simply arrange the events in chronological order as input to the predictor. Event chains contribute to improved overall performance, indicating that event chains provide more logically coherent event-level semantics, facilitating further reasoning and prediction. Together, these findings confirm that both detail-level and event-level semantics are essential for the success of VISTA in long-video event prediction.

\noindent \textbf{Effect of Specialized Components.}  We compare VISTA with the following three variants to examine the effectiveness of the key components in VISTA, as shown in Table~\ref{tab:one}:
(1) \textit{\textbf{w/o CVP}}: Without the character-centric visual prompt (CVP), the character portraits and raw video frames are directly feed into the VLM to generate visual descriptions. The marked performance degradation demonstrates the effectiveness of CVP in focusing the VLM on event-related visual details.
(2) \textit{\textbf{w/o KIR}}: Without the knowledge-enhanced iterative retrieval (KIR) strategy, all events are directly feed into the LLM to generate event chains. The noticeable performance degradation reflects the pivotal role KIR plays in enhancing LLM's casual analysis and reasoning over events.
(3) \textit{\textbf{w/o PtR}}: Without the propose-then-retrieve (PtR) strategy,  the LLM is asked to directly make prediction according to the input details and event chains. The evident performance drop indicates that PtR substantially enhances future-oriented analysis and clues integration during the prediction process.
Together, these results confirm that CVP, KIR, and PtR are indispensable components, jointly ensuring the robustness and effectiveness of VISTA in long-video event prediction.

\noindent
\begin{minipage}[t]{0.48\textwidth}
  \centering
  \captionof{table}{Ablation on multi-level semantics and components.}\label{tab:one}
  \begin{tabular}{c|c c c c | c}
        \hline
        \multicolumn{1}{c|}{} & R & M & C & SB & Acc \\
        \hline \hline
        VISTA  & 25.7 & 32.5 & 25.3 & 63.4& 62.8\\ 
        \hline
        w/o VD& 12.8 & 13.5 & 11.7 & 33.3 & 32.9\\
        w/o EC & 20.1 &23.7 & 18.2 & 49.1 & 49.2\\
        \hline
        w/o CVP & 23.7 &28.5 & 22.4 & 56.2 & 55.6\\ 
        w/o KIR & 20.4 & 24.0 & 18.5 & 51.3 & 51.9\\  
        w/o PtR & 19.8 & 25.5 & 19.7 & 54.0 & 52.9\\ 
        \hline
    \end{tabular}
\end{minipage}%
\hfill
\begin{minipage}[t]{0.48\textwidth}
  \centering
  \captionof{table}{Ablation on backbone LLM.}\label{tab:two}
  \vspace{0.45cm}
    \begin{tabular}{c|c c c c | c}
        \hline
        \multicolumn{1}{c|}{} & R & M & C & SB & Acc \\
        \hline \hline
        Llama3-8B & \underline{25.7} & \underline{32.5} & 25.3 & \underline{63.4} & \underline{62.8}\\ 
        Qwen2.5-7B & 24.9 & 32.6 & \underline{25.7} & 61.4 & 62.3\\  
        Mistral-7B & 24.3 & 31.9 & 24.1 & 60.7 & 61.1\\ 
        Gemma3-4B & 24.2 & 31.2 & 23.6 & 60.3 & 59.9\\ 
        \hline 
        GPT-4o  & 28.2 & 34.1 & 27.4 & 65.2& 66.5\\ 
        GPT-4.1 & \textbf{28.9} & \textbf{35.1} & \textbf{28.6} & \textbf{67.2}& \textbf{67.9}\\
        \hline
    \end{tabular}
\end{minipage}

\vspace{0.5cm}
\noindent \textbf{Effect of Backbone LLM.} To further analyze the generality of VISTA across different LLMs, we replace the default LLM in VISTA with the following: Qwen-2.5-7B-instruct, Mistral-7B-instruct-v0.2~\cite{jiang2024identifying},  Gemma3-4B~\cite{team2025gemma}, GPT-4o, and GPT-4.1. As shown in Table~\ref{tab:two}, the two commercial LLMs, GPT-4o and GPT-4.1, exhibit the best performance. Among the four smaller-scale open-source LLMs, LLaMA-3.1-8B-instruct (our default LLM ) performs best overall, while the other three deliver competitive results. Generally, it is evident that our VISTA demonstrates strong compatibility and scalability with various LLMs, proving generally effective for the widely-used LLMs.

\begin{figure}[H]
    \centering
    \includegraphics[width=\linewidth]{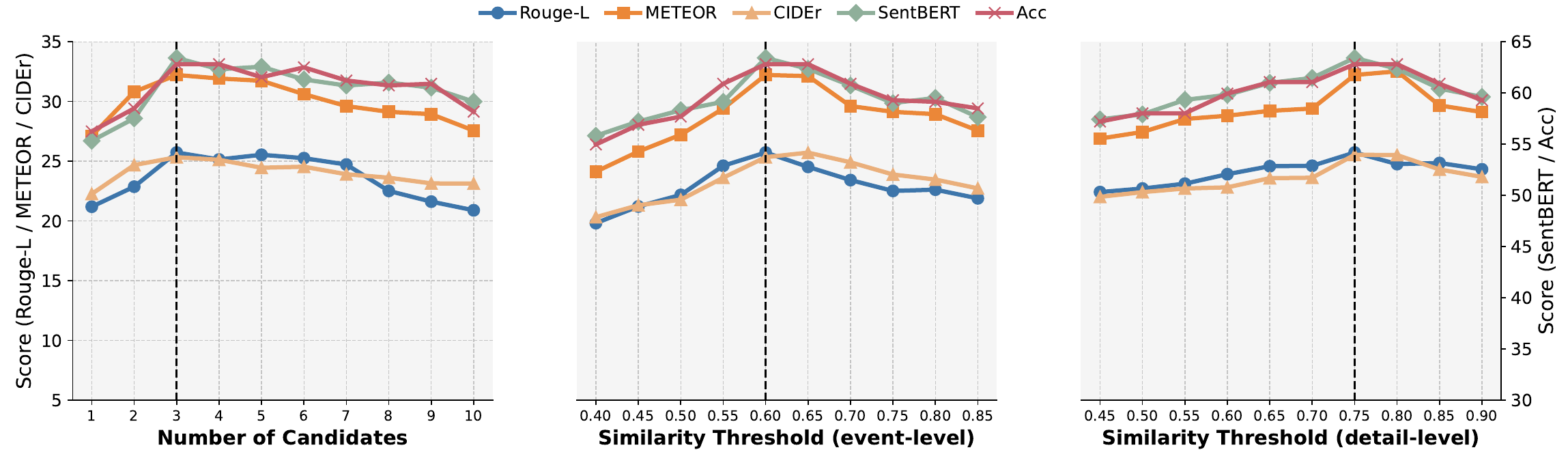}
    \caption{Ablation on three key hyperparamters of VISTA.}
    \label{figure:line}
\end{figure}

\noindent \textbf{Effect of Hyperparameters.}
As illustrated in Fig.\ref{figure:line}, we analyze the impact of three hyperparameters on VISTA’s performance: the candidate count $k$ in knowledge-enhanced iterative retrieval, and the similarity thresholds $\tau_e$ and $\tau_d$ in propose-then-retrieve. As these hyperparameters increase, VISTA’s performance curves exhibit similar rise–then–fall trends, peaking at $k=3$, $\tau_e=0.6$, and $\tau_d=0.75$. This is intuitively reasonable, as they all control the trade-off between diversity and interference in the retrieval process: higher $k$ or lower $\tau$ introduces more noise, while smaller $k$ or higher $\tau$ risks discarding valuable clues—both leading to degraded performance. Moreover, VISTA’s performance remains relatively stable across hyperparameter variations, with accuracy consistently above 50\%, demonstrating the robustness of the framework.
\vspace{-0.5cm}
\vspace{-1cm}

\subsection{Case Study}
\begin{figure}
    \centering
    \includegraphics[width=\linewidth]{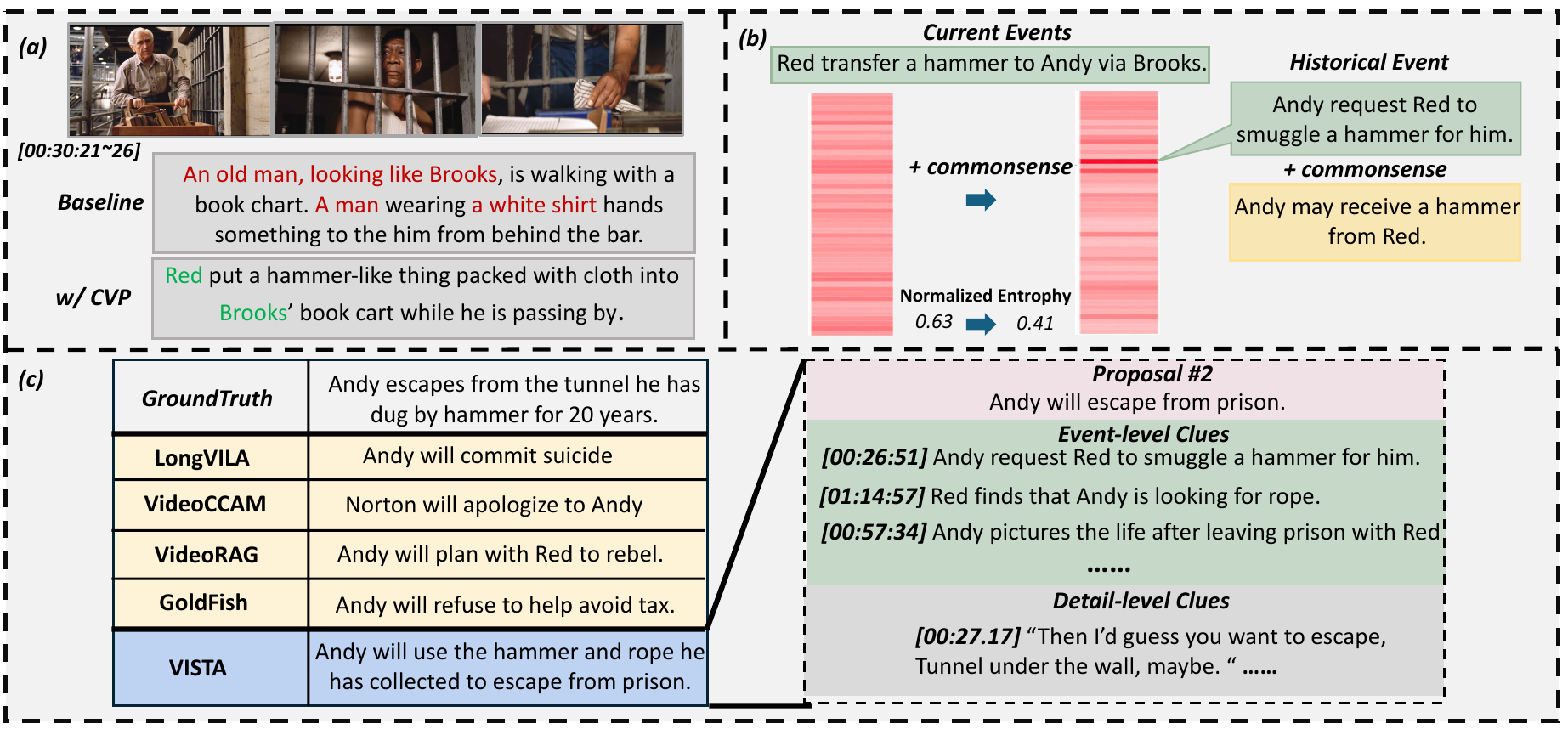}
    \caption{Visualization of a long-video event prediction case.}
    \label{fig:case}
\end{figure}
As shown in Fig.\ref{fig:case}, we present visualization results to qualitatively analyze VISTA and its internal components. 

Part (a) shows the VLM output with and without the character-centric visual prompt(CVP). With CVP, the model accurately identifies the characters \textit{Red} and \textit{Brooks} and concisely describes their actions. In contrast, the baseline not only fails to accurately identify their identities but also includes trivial event-irrelevant details such as \textit{white shirt}.

Part(b) shows heatmaps of semantic similarity between embeddings of historical and current events, where darker red indicates higher similarity. We compute the normlized entrophy(NE) to judge the distribution evenness, where lower NE represents more uneven distribution. It is evident that the distribution of post-commonsense heatmap become more uneven after incorporating commonsense, with the similarity between the two causally-related events - \textit{"Andy request Red to smuggle a hammer for him"} and current event - significantly increased. This proves the effectiveness of commonsense expert to facilitate more accurate causal reasoning during event chain construction. 

Part(c) shows the predictions of various methods on a sample from \textit{The Shawshank Redemption}. We find that all general LVLMs’ predictions deviate significantly from the ground truth, whereas VISTA not only aligns with the ground truth but also contains concrete details (\textit{using the hammer and rope}), showing its superiority in long video event prediction. We further extract VISTA's intermediate results and find that during the propose-then-retrieve, VISTA first generates a proposal \textit{"Andy will escape from prison"},  and then collects key clues such as Andy's prior acquisition of tools (\textit{hammer} and \textit{rope}), as well as Red's suspicion when Andy requested the hammer. These clues makes VISTA prioritizes this proposal and further refine it with collected clues into the final prediction. This both demonstrates the effectiveness of the propose-then-retrieve strategy and highlights the explainability of our VISTA framework.

\section{Conclusion}
In this paper, we propose VISTA, a multi-level event semantics mining framework for long-video event prediction. VISTA first extracts precise event-related visual details by the character-centric visual prompt, then progressively constructs structured and coherent event chains through a knowledge-enhanced iterative retrieval strategy to capture event evolution. Finally, it adopts a human-like propose-then-retrieve strategy, which generates diverse predictive proposals from event chains and refines them by retrieving detail- and event-level clues from historical information, ultimately producing accurate and explainable predictions. Extensive experiments on real-world datasets demonstrate the superiority of VISTA, while ablation and qualitative studies further validate the robustness and effectiveness of its internal design.

\noindent\textbf{Acknowledgements.} This work was supported by the grants from National Natural Science Foundation of China (No.62222213, 62072423).
\newpage
\bibliographystyle{splncs04}
\bibliography{refs}

@article{lei2020more,
  title={What is more likely to happen next? video-and-language future event prediction},
  author={Lei, Jie and Yu, Licheng and Berg, Tamara L and Bansal, Mohit},
  journal={arXiv preprint arXiv:2010.07999},
  year={2020}
}

@inproceedings{song2024moviechat,
  title={Moviechat: From dense token to sparse memory for long video understanding},
  author={Song, Enxin and Chai, Wenhao and Wang, Guanhong and Zhang, Yucheng and Zhou, Haoyang and Wu, Feiyang and Chi, Haozhe and Guo, Xun and Ye, Tian and Zhang, Yanting and others},
  booktitle={Proceedings of the IEEE/CVF Conference on Computer Vision and Pattern Recognition},
  pages={18221--18232},
  year={2024}
}

@inproceedings{fan2025videoagent,
  title={Videoagent: A memory-augmented multimodal agent for video understanding},
  author={Fan, Yue and Ma, Xiaojian and Wu, Rujie and Du, Yuntao and Li, Jiaqi and Gao, Zhi and Li, Qing},
  booktitle={European Conference on Computer Vision},
  pages={75--92},
  year={2025},
  organization={Springer}
}

@inproceedings{li2025llama,
  title={Llama-vid: An image is worth 2 tokens in large language models},
  author={Li, Yanwei and Wang, Chengyao and Jia, Jiaya},
  booktitle={European Conference on Computer Vision},
  pages={323--340},
  year={2025},
  organization={Springer}
}

@inproceedings{shtedritski2023does,
  title={What does clip know about a red circle? visual prompt engineering for vlms},
  author={Shtedritski, Aleksandar and Rupprecht, Christian and Vedaldi, Andrea},
  booktitle={Proceedings of the IEEE/CVF International Conference on Computer Vision},
  pages={11987--11997},
  year={2023}
}

@article{song2024moviellm,
  title={Moviellm: Enhancing long video understanding with ai-generated movies},
  author={Song, Zhende and Wang, Chenchen and Sheng, Jiamu and Zhang, Chi and Yu, Gang and Fan, Jiayuan and Chen, Tao},
  journal={arXiv preprint arXiv:2403.01422},
  year={2024}
}

@inproceedings{wang2025videoagent,
  title={Videoagent: Long-form video understanding with large language model as agent},
  author={Wang, Xiaohan and Zhang, Yuhui and Zohar, Orr and Yeung-Levy, Serena},
  booktitle={European Conference on Computer Vision},
  pages={58--76},
  year={2025},
  organization={Springer}
}

@inproceedings{vedantam2015cider,
  title={Cider: Consensus-based image description evaluation},
  author={Vedantam, Ramakrishna and Lawrence Zitnick, C and Parikh, Devi},
  booktitle={Proceedings of the IEEE conference on computer vision and pattern recognition},
  pages={4566--4575},
  year={2015}
}

@inproceedings{banerjee2005meteor,
  title={METEOR: An automatic metric for MT evaluation with improved correlation with human judgments},
  author={Banerjee, Satanjeev and Lavie, Alon},
  booktitle={Proceedings of the acl workshop on intrinsic and extrinsic evaluation measures for machine translation and/or summarization},
  pages={65--72},
  year={2005}
}

@inproceedings{lin2004rouge,
  title={Rouge: A package for automatic evaluation of summaries},
  author={Lin, Chin-Yew},
  booktitle={Text summarization branches out},
  pages={74--81},
  year={2004}
}

@article{team2025gemma,
  title={Gemma 3 technical report},
  author={Team, Gemma and Kamath, Aishwarya and Ferret, Johan and Pathak, Shreya and Vieillard, Nino and Merhej, Ramona and Perrin, Sarah and Matejovicova, Tatiana and Ram{\'e}, Alexandre and Rivi{\`e}re, Morgane and others},
  journal={arXiv preprint arXiv:2503.19786},
  year={2025}
}

@article{halawi2024approaching,
  title={Approaching human-level forecasting with language models},
  author={Halawi, Danny and Zhang, Fred and Yueh-Han, Chen and Steinhardt, Jacob},
  journal={arXiv preprint arXiv:2402.18563},
  year={2024}
}

@article{luo2024video,
  title={Video-RAG: Visually-aligned Retrieval-Augmented Long Video Comprehension},
  author={Luo, Yongdong and Zheng, Xiawu and Yang, Xiao and Li, Guilin and Lin, Haojia and Huang, Jinfa and Ji, Jiayi and Chao, Fei and Luo, Jiebo and Ji, Rongrong},
  journal={arXiv preprint arXiv:2411.13093},
  year={2024}
}

@inproceedings{ma2023context,
  title={Context-aware event forecasting via graph disentanglement},
  author={Ma, Yunshan and Ye, Chenchen and Wu, Zijian and Wang, Xiang and Cao, Yixin and Chua, Tat-Seng},
  booktitle={Proceedings of the 29th ACM SIGKDD Conference on Knowledge Discovery and Data Mining},
  pages={1643--1652},
  year={2023}
}

@article{chen2024longvila,
  title={Longvila: Scaling long-context visual language models for long videos},
  author={Chen, Yukang and Xue, Fuzhao and Li, Dacheng and Hu, Qinghao and Zhu, Ligeng and Li, Xiuyu and Fang, Yunhao and Tang, Haotian and Yang, Shang and Liu, Zhijian and others},
  journal={arXiv preprint arXiv:2408.10188},
  year={2024}
}

@article{grattafiori2024llama,
  title={The llama 3 herd of models},
  author={Grattafiori, Aaron and Dubey, Abhimanyu and Jauhri, Abhinav and Pandey, Abhinav and Kadian, Abhishek and Al-Dahle, Ahmad and Letman, Aiesha and Mathur, Akhil and Schelten, Alan and Vaughan, Alex and others},
  journal={arXiv e-prints},
  pages={arXiv--2407},
  year={2024}
}

@article{fei2024video,
  title={Video-ccam: Enhancing video-language understanding with causal cross-attention masks for short and long videos},
  author={Fei, Jiajun and Li, Dian and Deng, Zhidong and Wang, Zekun and Liu, Gang and Wang, Hui},
  journal={arXiv preprint arXiv:2408.14023},
  year={2024}
}

@article{lai2024object,
  title={Object-Centric Cross-Modal Knowledge Reasoning for Future Event Prediction in Videos},
  author={Lai, Chenghang and Wang, Haibo and Ge, Weifeng and Xue, Xiangyang},
  journal={IEEE Transactions on Circuits and Systems for Video Technology},
  year={2024},
  publisher={IEEE}
}

@article{bain2023whisperx,
  title={Whisperx: Time-accurate speech transcription of long-form audio},
  author={Bain, Max and Huh, Jaesung and Han, Tengda and Zisserman, Andrew},
  journal={arXiv preprint arXiv:2303.00747},
  year={2023}
}

@article{ghermi2024short,
  title={Short film dataset (sfd): A benchmark for story-level video understanding},
  author={Ghermi, Ridouane and Wang, Xi and Kalogeiton, Vicky and Laptev, Ivan},
  journal={arXiv preprint arXiv:2406.10221},
  year={2024}
}

@inproceedings{bain2020condensed,
  title={Condensed movies: Story based retrieval with contextual embeddings},
  author={Bain, Max and Nagrani, Arsha and Brown, Andrew and Zisserman, Andrew},
  booktitle={Proceedings of the Asian Conference on Computer Vision},
  year={2020}
}

@inproceedings{sap2019atomic,
  title={Atomic: An atlas of machine commonsense for if-then reasoning},
  author={Sap, Maarten and Le Bras, Ronan and Allaway, Emily and Bhagavatula, Chandra and Lourie, Nicholas and Rashkin, Hannah and Roof, Brendan and Smith, Noah A and Choi, Yejin},
  booktitle={Proceedings of the AAAI conference on artificial intelligence},
  volume={33},
  pages={3027--3035},
  year={2019}
}

@article{li2021future,
  title={The future is not one-dimensional: Complex event schema induction by graph modeling for event prediction},
  author={Li, Manling and Li, Sha and Wang, Zhenhailong and Huang, Lifu and Cho, Kyunghyun and Ji, Heng and Han, Jiawei and Voss, Clare},
  journal={arXiv preprint arXiv:2104.06344},
  year={2021}
}

@article{lyu2024generating,
  title={Generating Event-oriented Attribution for Movies via Two-Stage Prefix-Enhanced Multimodal LLM},
  author={Lyu, Yuanjie and Xu, Tong and Niu, Zihan and Peng, Bo and Ke, Jing and Chen, Enhong},
  journal={arXiv preprint arXiv:2409.09362},
  year={2024}
}

@article{reimers2019sentence,
  title={Sentence-BERT: Sentence Embeddings using Siamese BERT-Networks},
  author={Reimers, N},
  journal={arXiv preprint arXiv:1908.10084},
  year={2019}
}

@article{ataallah2024goldfish,
  title={Goldfish: Vision-language understanding of arbitrarily long videos},
  author={Ataallah, Kirolos and Shen, Xiaoqian and Abdelrahman, Eslam and Sleiman, Essam and Zhuge, Mingchen and Ding, Jian and Zhu, Deyao and Schmidhuber, J{\"u}rgen and Elhoseiny, Mohamed},
  journal={arXiv preprint arXiv:2407.12679},
  year={2024}
}

@article{he2024storyteller,
  title={StoryTeller: Improving Long Video Description through Global Audio-Visual Character Identification},
  author={He, Yichen and Lin, Yuan and Wu, Jianchao and Zhang, Hanchong and Zhang, Yuchen and Le, Ruicheng},
  journal={arXiv preprint arXiv:2411.07076},
  year={2024}
}

@article{chung2024scaling,
  title={Scaling instruction-finetuned language models},
  author={Chung, Hyung Won and Hou, Le and Longpre, Shayne and Zoph, Barret and Tay, Yi and Fedus, William and Li, Yunxuan and Wang, Xuezhi and Dehghani, Mostafa and Brahma, Siddhartha and others},
  journal={Journal of Machine Learning Research},
  volume={25},
  number={70},
  pages={1--53},
  year={2024}
}

@mastersthesis{jiang2024identifying,
  title={Identifying and mitigating vulnerabilities in llm-integrated applications},
  author={Jiang, Fengqing},
  year={2024},
  school={University of Washington}
}

@inproceedings{ren2023pbidr,
  title={Facial Geometric Detail Recovery via Implicit Representation},
  author={Ren, Xingyu and Lattas, Alexandros and Gecer, Baris and Deng, Jiankang and Ma, Chao and Yang, Xiaokang},
  booktitle={2023 IEEE 17th International Conference on Automatic Face and Gesture Recognition (FG)},  
  year={2023}
 }

@inproceedings{radford2021learning,
  title={Learning transferable visual models from natural language supervision},
  author={Radford, Alec and Kim, Jong Wook and Hallacy, Chris and Ramesh, Aditya and Goh, Gabriel and Agarwal, Sandhini and Sastry, Girish and Askell, Amanda and Mishkin, Pamela and Clark, Jack and others},
  booktitle={International conference on machine learning},
  pages={8748--8763},
  year={2021},
  organization={PmLR}
}

@inproceedings{shu2025video,
  title={Video-xl: Extra-long vision language model for hour-scale video understanding},
  author={Shu, Yan and Liu, Zheng and Zhang, Peitian and Qin, Minghao and Zhou, Junjie and Liang, Zhengyang and Huang, Tiejun and Zhao, Bo},
  booktitle={Proceedings of the Computer Vision and Pattern Recognition Conference},
  pages={26160--26169},
  year={2025}
}

@article{wang2024longllava,
  title={Longllava: Scaling multi-modal llms to 1000 images efficiently via a hybrid architecture},
  author={Wang, Xidong and Song, Dingjie and Chen, Shunian and Zhang, Chen and Wang, Benyou},
  journal={arXiv preprint arXiv:2409.02889},
  year={2024}
}

\end{document}